\documentclass{article}

\usepackage{arxiv}

\usepackage[utf8]{inputenc} 
\usepackage[T1]{fontenc}    
\usepackage{hyperref}       
\usepackage{url}            
\usepackage{booktabs}       
\usepackage{amsfonts}       
\usepackage{nicefrac}       
\usepackage{microtype}      
\usepackage{lipsum}
\usepackage{graphicx}
\usepackage{epstopdf}
\usepackage{amsmath,amssymb}
\usepackage{bm}
\usepackage{subcaption}
\usepackage{xcolor}
\hypersetup{
  colorlinks,
  linkcolor={blue!50!black},
  citecolor={blue!50!black},
  urlcolor={blue!80!black},
  anchorcolor = {blue!80!black},
  filecolor = {blue!80!black},
  menucolor = {blue!80!black},
  runcolor = {blue!80!black}
}
\newcommand\blfootnote[1]{%
  \begingroup
  \renewcommand\thefootnote{}\footnote{#1}%
  \addtocounter{footnote}{-1}%
  \endgroup
}

\title{Gradient-enhanced multifidelity neural networks for high-dimensional function approximation}

\author{
 Jethro Nagawkar \\
  Department of Aerospace Engineering\\
  Iowa State University\\
  Ames, Iowa, 50011 \\
  \texttt{jethro@iastate.edu} \\
   \And
 Leifur Leifsson \\
  Department of Aerospace Engineering\\
  Iowa State University\\
  Ames, Iowa, 50011 \\
  \texttt{leifur@iastate.edu} \\
}

\begin{document}
\maketitle
\begin{abstract}
In this work, a novel multifidelity machine learning (ML) model, the gradient-enhanced multifidelity neural networks (GEMFNNs), is proposed. This model is a multifidelity version of gradient-enhanced neural networks (GENNs) as it uses both function and gradient information available at multiple levels of fidelity to make function approximations. Its construction is similar to multifidelity neural networks (MFNNs). This model is tested on three analytical function, a one, two, and a 20 variable function. It is also compared to neural networks (NNs), GENNs, and MFNNs, and the number of samples required to reach a global accuracy of 0.99 coefficient of determination ($R^2$) is measured. GEMFNNs required 18, 120, and 600 high-fidelity samples for the one, two, and 20 dimensional cases, respectively, to meet the target accuracy. NNs performed best on the one variable case, requiring only ten samples, while GENNs worked best on the two variable case, requiring 120 samples. GEMFNNs worked best for the 20 variable case, while requiring nearly eight times fewer samples than its nearest competitor, GENNs. For this case, NNs and MFNNs did not reach the target global accuracy even after using 10,000 high-fidelity samples. This work demonstrates the benefits of using gradient as well as multifidelity information in NNs for high-dimensional problems.\blfootnote{Preprint submitted to AMSE 2021 IDETC/CIE conference}
\end{abstract}


\section{Introduction}
Surrogate modeling methods are useful for reducing the computational cost in various problems involving optimum design \cite{Li2019, jethro2021}, uncertainty quantification \cite{Jethro2020,du2019b}, and global sensitivity analysis \cite{Wang2013,du2019}. In these methods, a computationally efficient surrogate model replaces an expensive physics-based model, reducing the overall cost involved in solving such problems. 

Surrogate modeling methods can be broadly classified as either being data-fit methods \cite{Queipo2005} or multifidelity methods \cite{Peherstorfer2018}. In data-fit methods, a response surface is fitted through evaluated single-fidelity sample points. While in multifidelity methods, low-fidelity data is used to augment the predictive capabilities of surrogate models constructed from a limited number of high-fidelity data. High-fidelity models are models that solve the task at hand with a desired accuracy. Low-fidelity models solve the same task but at lower cost and with lower accuracy.

A variety data-fit methods exist in literature such as Kriging \cite{Krige1951} (also known as Gaussian process regression) and its variants \cite{Schobi2015, Bouhlel2016, Bouhlel2016b}, polynomial chaos expansions \cite{Blatman2009}, and support vector machines \cite{Li2015}. Of these methods, Kriging is the most widely used method in various engineering analysis and design tasks \cite{Forrester2008}. Kriging, however, suffers from a variety of issues such as being poor at approximating discontinuous functions \cite{raissi2016deep}, difficulty in handling high-dimensional problems \cite{Perdikaris2017}, costly to use in the presence of a large number of data samples \cite{Forrester2008}, and being difficult to implement \cite{Meng2020}. Several methods have been introduced to handle these issues such as the use of gradient information \cite{Forrester2008, Bouhlel2019}, and the partial least-squares correlation functions \cite{Bouhlel2016,Bouhlel2016b}. While improvements have been reported, several of the key issues still remain \cite{Meng2020,Bouhlel2020}. 

Multifidelity methods have been introduced as a way of reducing the overall cost involved in engineering design and analysis tasks \cite{Forrester2008, Peherstorfer2018}. The key benefit is the use of low-fidelity data, along with a limited number of high-fidelity data, reducing the overall cost in acquiring the data to construct the surrogate model. Cokriging \cite{Kennedy2000}, the multifidelity version of Kriging, and its variants \cite{Du2020,Deng2020}, while becoming popular in design and analysis tasks, still suffers from the issues associated with Kriging \cite{Meng2020}. 

The use of neural networks (NNs) \cite{goodfellow2016} in engineering design and analysis problems is becoming more prevalent \cite{Meng2020,Bouhlel2020}. NNs overcome many of the major challenges in Kriging models such as the ability to handle high-dimensional datasets \cite{Bouhlel2020}, scalability with the number of data \cite{goodfellow2016}, easier to implement \cite{Meng2020}, as well as being good at approximating discontinuous data \cite{Meng2020}. The major drawback of NNs is that they require a large number of samples to make accurate predictions \cite{goodfellow2016}, especially for high-dimensional problems \cite{Bouhlel2020}. To overcome this challenge, different NN variants, such as gradient-enhanced NNs (GENNs) \cite{Bouhlel2020}, and multifidelity NNs (MFNNs) \cite{Meng2020}, have been recently introduced. 

In this work, a multifidelity variant of GENNs \cite{Bouhlel2020}, gradient-enhanced MFNNs (GEMFNNs) is introduced and demonstrated on three different analytical problems, involving one, two, and 20 variables. The proposed approach is compared to NNs, GENNs, and MFNNs for these problems. GEMFNNs are constructed in a similar fashion as MFNNs by leveraging both function and gradient information available from low- and high-fidelity models to yield accurate function approximations. To the author's knowledge, the proposed GEMFNNs is a novel machine learning (ML) modeling algorithm. 

The remainder of this paper is organized in the following way. The next section describes the methods used to construct the GEMFNNs ML model. In the following section, the GEMFNNs model is demonstrated on three analytical benchmark problems. This paper then ends with the conclusion and future work.

\section{Methods}
This section describes the construction of the GEMFNNs ML model. An outline of the GEMFNNs-based analysis is first introduced, followed by the sampling plan used to generate the data, which is needed in order to train and test this ML model. The construction methodology of the GEMFNNs is discussed in the following section, followed by the validation metric used to quantify its global accuracy. Finally, this model can be used for further analysis, such as optimal design and uncertainty quantification. 

\subsection{Outline of the GEMFNNs Construction}
\begin{figure}[t]
\begin{center}
\includegraphics[trim={0.5cm 2.5cm 0.5cm 2.5cm},clip,width=0.75\textwidth]{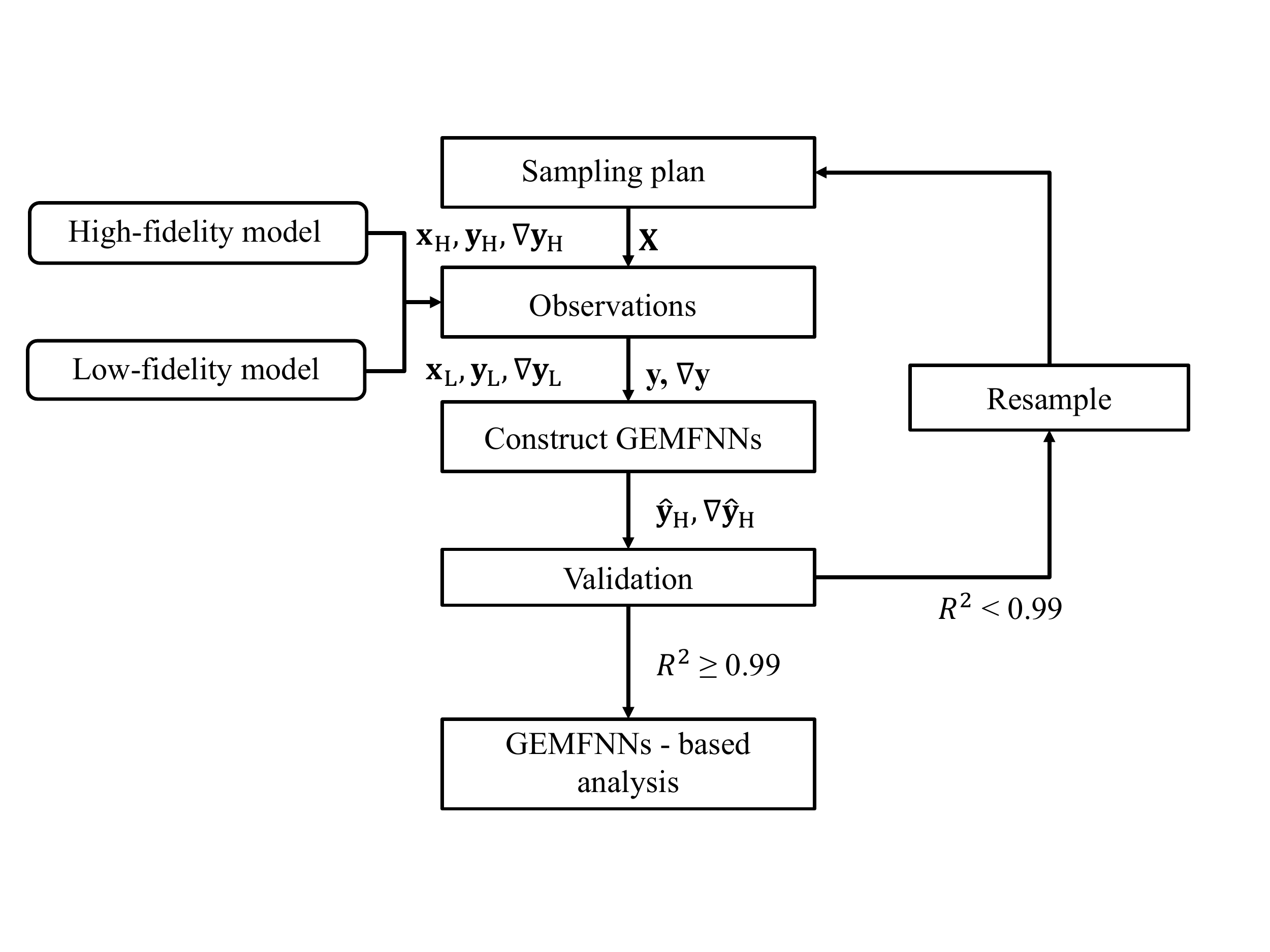}
\end{center}
\caption{Flowchart of the gradient-enhanced multifidelity neural network construction.}
\label{flow} 
\end{figure}

A flowchart of the GEMFNN construction is shown in Fig.~\ref{flow}. It begins by sampling the input design space, $\textbf{X} \in \mathbb{R}^{m\times n}$, first, in order to generate the data required to both train and test the ML model. $m$ is the number of samples and $n$ is the number of input variables. The training data consists of two different sample sets, one for evaluating the high-fidelity model, $\textbf{x}_\text{H} \in \mathbb{R}^{m_\text{H}\times n}$, and the other for evaluating the low-fidelity model, $\textbf{x}_\text{L} \in \mathbb{R}^{m_\text{L}\times n}$. $m_\text{H}$ and $m_\text{L}$ are the number of high- and low-fidelity samples, respectively. Both the function and its gradients need to be evaluated at the high- ($\textbf{y}_\text{H} \in \mathbb{R}^{m_\text{H}\times 1}$ and $\nabla \textbf{y}_\text{H} \in \mathbb{R}^{m_\text{H}\times n}$) and low-fidelity models ($\textbf{y}_\text{L} \in \mathbb{R}^{m_\text{L}\times 1}$ and $\nabla \textbf{y}_\text{L} \in \mathbb{R}^{m_\text{L}\times n}$), respectively. $\textbf{y}_\text{L}$ and $\textbf{y}_\text{H}$ together represent the combined observation $\textbf{y} \in \mathbb{R}^{m\times 1}$, while $\nabla \textbf{y}_\text{L}$ and $\nabla \textbf{y}_\text{H}$ together represent the combined observation $\nabla \textbf{y} \in \mathbb{R}^{m\times n}$. A separate testing set is created by evaluating only the high-fidelity model's function and its gradients. $\hat{\textbf{y}}_\text{H} \in \mathbb{R}^{m_\text{H}\times 1}$ and $\nabla \hat{\textbf{y}}_\text{H} \in \mathbb{R}^{m_\text{H}\times n}$ are the high-fidelity function and gradient predictions, respectively, from the GEMFNNs. The accuracy of these predictions are then measured using the coefficient of determination ($R^2$) error metric. The above process is repeated several times, each with an increasing training sample size, until terminating on the validation criteria. On terminating, GEMFNNs can be used for further analysis in engineering design and analysis.

\subsection{Sampling Plan}
Sampling is the first process involved in constructing the GEMFNNs. It is the process of selecting discrete samples in the variable space \cite{Forrester2008}. In this study , both the full factorial sampling plan \cite{Forrester2008}, as well as the Latin Hypercube sampling (LHS) \cite{mckay1979} plan are used to generate both the training and testing data. The choice of the sampling plan used is case dependent and is discussed in their corresponding sections. 

\begin{figure}[b]
\begin{center}
\includegraphics[trim={4.5cm 4.2cm 3.5cm 2.5cm},clip,width=0.75\textwidth]{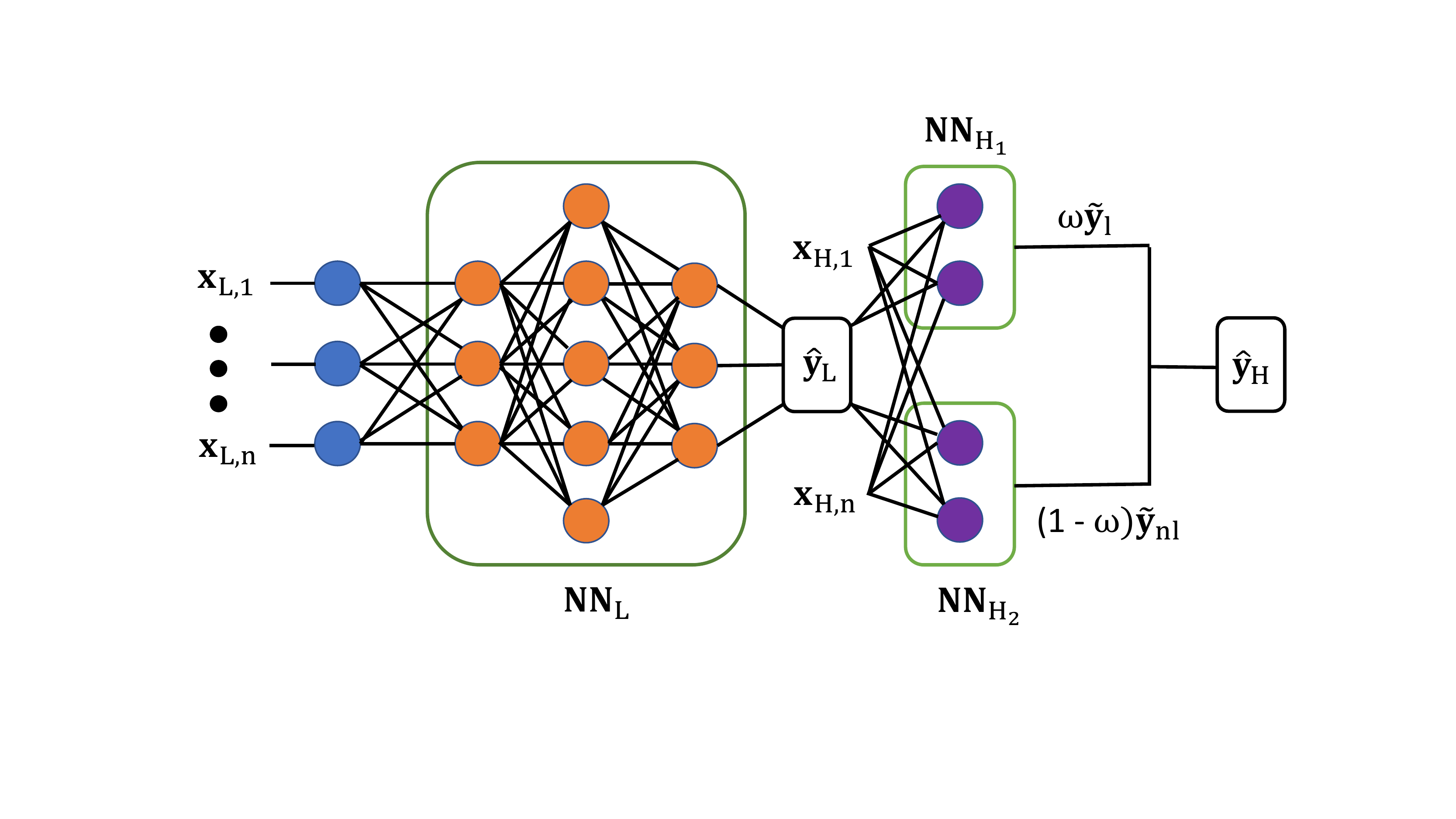}
\end{center}
\caption{Multifidelity neural network architecture.}
\label{arch} 
\end{figure}

\subsection{Gradient-Enhanced Multifidelity Neural Networks}

NNs are universal function approximators \cite{goodfellow2016}, where a hierarchy of features, known as layers, is used to approximated any given function. The layers in-between the input and output layers are called hidden layers. The output and hidden layers contain neurons, which are a fundamental unit of computation and contain an activation function \cite{goodfellow2016}. In NNs, an unconstrained optimization problem is solved, where the parameters of the NNs are tuned using the Adaptive Moments (ADAM) \cite{kingma2014} gradient-based optimizer \cite{goodfellow2016}, where the backpropagation algorithm \cite{chauvin1995} is used to compute the gradients. In this study, the mean squared error (MSE) is used as the loss function in the optimization problem is given by

\begin{equation}\label{nn_loss}
    \mathcal{L}_{NN} = \frac{\sum\nolimits_{l=1}^{N}  (\hat{y}_\text{H}^{(l)} - y_\text{H}^{(l)})^2}{N},
\end{equation}
where $N$ the number of samples in a subset of the training data, called mini-batch \cite{goodfellow2016}, is used to minimize the mismatch between the high-fidelity training data observations, $y_\text{H}$, and the predicted values, $\hat{y}_\text{H}$, of the NN.

GENNs \cite{Bouhlel2020} modify the loss function in (\ref{nn_loss}) by adding the mismatch match between the high-fidelity training data gradient, $\nabla y_\text{H}$, and the predicted gradient of the NNs, $\nabla \hat{y}
_\text{H}$, to it, and is given by 
\begin{equation}\label{genn_loss}
    \mathcal{L}_{GENN} = \mathcal{L}_{NN} + \frac{\sum\nolimits_{l=1}^{N} \sum\nolimits_{k=1}^{D}  (\nabla \hat{y}_{\text{H},k}^{(l)} - \nabla y_{\text{H},k}^{(l)})^2}{N},
\end{equation}
where $D$ is the dimension of the input variable space. This loss function ensures a reduction in the mismatch between both the function and its tangent at a given training point to the corresponding true values, respectively.  

A schematic of the MFNN architecture is shown in Fig. \ref{arch}. It contains three NNs, $NN_L$, which is used to approximate the low-fidelity data, the output of which is used as an additional input variable to two other NNs, $NN_{H_1}$ and $NN_{H_2}$. $NN_{H_1}$ and $NN_{H_2}$ are used to capture the linear ($\Tilde{y}_l$) and nonlinear correlations ($\Tilde{y}_{nl}$) between high- and low-fidelity data. $NN_{H_1}$ contains linear activation functions, while $NN_{H_2}$ contains nonlinear activation functions. The weighted sum of the outputs of the linear and nonlinear layer gives the high-fidelity prediction of the MFNNs as
\begin{equation}
    \hat{y}\textsubscript{H} = \omega\Tilde{y}_l + (1-\omega)\Tilde{y}_{nl},
\end{equation}
where $\omega$ is an additional parameter of the MFNNs. Note that NNs and GENNs use only the $NN_{H_2}$ part of MFNNs, but do not include $\hat{y}_\text{L}$ as an additional input parameter. The loss function of MFNNs is given as

\begin{equation}\label{mfnn_loss}
    \mathcal{L}_{MFNN} = \mathcal{L}_{NN} + \frac{\sum\nolimits_{l=1}^{N}  (\hat{y}_\text{L}^{(l)} - y_\text{L}^{(l)})^2}{N}.
\end{equation}

GEMFNNs is a multifidelity version of GENNs and is constructed similar to MFNNs. It uses gradient information available at high and low-fidelity data during training. The loss function for GEMFNNs is taken as 

\begin{equation}\label{gemfnn_loss}
\begin{aligned}
    \mathcal{L}_{GEMFNN} = &\mathcal{L}_{MFNN} + \frac{\sum\nolimits_{l=1}^{N} \sum\nolimits_{k=1}^{D} (\nabla \hat{y}_{\text{L},k}^{(l)} - \nabla y_{\text{L},k}^{(l)})^2}{N}\\  &+ \frac{\sum\nolimits_{l=1}^{N} \sum\nolimits_{k=1}^{D} (\nabla \hat{y}_{\text{H},k}^{(l)} - \nabla y_{\text{H},k}^{(l)})^2}{N}.
\end{aligned}
\end{equation}

The steps involved in the construction of GEMFNNs are as follows:
\begin{enumerate}
    \item Normalize the input, output and gradient of output with respect the inputs for all the data.
    \item Perform forward propagation through $NN_L$ to get $\hat{y}
    _\text{L}$. Then do the same through the MFNNs to get $\hat{y}_\text{H}$. 
    \item Use reverse mode automatic differentiation \cite{Rall1981} to calculate both $\nabla \hat{y}_\text{L}$ and $\nabla \hat{y}_\text{H}$. 
    \item Calculate $\mathcal{L}_{GEMFNN}$ using (\ref{gemfnn_loss}).
    \item Use backpropagation \cite{chauvin1995} to calculate the gradient of $\mathcal{L}_{GEMFNN}$ with respect to all the parameters in GEMFNNs ($\mathbf{\theta}$), given by $\nabla \mathcal{L}_{GEMFNN}$.
    \item Update the parameters:
    \begin{equation}
    \mathbf{\theta} \leftarrow \mathbf{\theta} - \alpha \nabla \mathcal{L}_{GEMFNN},
    \end{equation}
    where $\alpha$ is the learning rate hyperparameter.
    \item Iterate over steps $2-6$ till all the mini-batches present in one epoch is used. One epoch refers to one iteration over an entire training dataset \cite{goodfellow2016}. 
    \item Repeat steps $2-7$ for all the epochs. 
    \item GEMFNNs is now trained and ready to be used for function approximation. 
\end{enumerate}

\subsection{Validation}

In this work, the coefficient of determination is used to measure the global accuracy of the ML models, $R^2$, given as
\begin{equation}
    \text{$R^2$} = 1 - \frac{\sum\nolimits_{j=1}^{N_{t}} (y^{(j)}_{t} - \hat{y}^{(j)}_{t})^2}{\sum\nolimits_{j=1}^{N_{t}} (y^{(j)}_{t} - \bar{y}_{t})^2},
\end{equation}
where $N_t$ is the total number of testing data samples, $\hat{y}^{(j)}_{t}$ and $y^{(j)}_{t}$ are the ML model estimation and high-fidelity observation of the $j^{\text{th}}$ testing point, respectively, and $\bar{y}_{t}$ is the mean of $y^{(j)}_{t}$, given by
\begin{equation}
    \bar{y}_{t} = \frac{\sum\nolimits_{j=1}^{N_{t}} y^{(j)}_{t}}{N_{t}}.
\end{equation}
$R^2$ is the measure of ``Goodness of fit'' \cite{Forrester2008} of a model. When R$^2$ equals one, the model has approximated the true function perfectly. This makes it easier to chose the global accuracy criterion. In this work, a value of $R^2$ greater than $0.99$ is considered an acceptable global accuracy.

In this work, for each high-fidelity sample size and for each ML model, the mean and the standard deviation of the $R^2$ metric is plotted using `$n_t$' different datasets. This is done in order to account for the variation in the training data used as well as due to the stochastic nature of the ADAM optimizer \cite{kingma2014}. The mean of $R^2$ is
\begin{equation}
    \mu_\text{$R^2$} = \frac{\sum\nolimits_{k=1}^{n_t} \text{$R_k^2$}} {n_t},
\end{equation}
and the standard deviation is 
\begin{equation}
    \sigma_\text{$R^2$} = \sqrt{\frac{\sum\nolimits_{k=1}^{n_t} (\text{$R_k^2$}-\mu_\text{$R^2$})^2} {n_t}}.
\end{equation}
In this work, $n_t$ is set to ten for all the cases.

\section{Numerical examples}
In this study, the GEMFNNs ML model is demonstrated on three different analytical problems. The first is an one variable analytical function. The second, the two variable Rastrigin function \cite{Rastrigin1974}, and the final a 20 variable analytical function. The GEMFNNs ML model is compared to other ML models, namely, NNs, GENNs, and MFNNs.

\subsection{Case 1: one dimensional analytical function}

The one dimensional analytical function used in this study was developed by Forrester et al. \cite{Forrester2008} and is written as
\begin{equation}
    f_{\text{HF}}(\textbf{x}) = (6x-2)^2sin(12x-4),
\end{equation}
where $x$ $\in$ [0,1]. Forrester et al. \cite{Forrester2008} also introduced a corresponding low-fidelity model given by 

\begin{equation}
    f_{\text{LF}}(\textbf{x}) = 0.5f_{\text{HF}}(\textbf{x}) + 10(x-0.5) -5.
\end{equation}
The corresponding gradient of the high-fidelity model is 
\begin{equation}
\begin{aligned}
    \nabla f_{\text{HF}}(\textbf{x}) = &12(6x-2)sin(12x-4) \\&+ 12(6x-2)^2cos(12x-4),
\end{aligned}
\end{equation}
and the low-fidelity model gradient is
\begin{equation}
    \nabla f_{\text{LF}}(\textbf{x}) = 0.5\nabla f_{\text{HF}}(\textbf{x}) + 10.
\end{equation}

\subsubsection{ML model setup}
In this case, the full factorial sampling plan is used to generate both the high- and low-fidelity training data, as well as the high-fidelity testing data. The testing data contains $1,000$ samples. As discussed the in the previous section, NNs and GENNs use only the $NN_{H_2}$ section of the composite NN, while MFNNs and GEMFNNs use $NN_{L}$, $NN_{H_1}$, and $NN_{H_2}$. Note that the various hyperparameters used for this case are kept the same for the different ML models. $NN_{H_1}$ and $NN_{H_2}$ for this case is set to include only one hidden layer, with ten neurons, while $NN_{L}$ has 20 neurons and one hidden layer. Both $NN_{L}$ and $NN_{H_2}$ use the tangent hyperbolic activation function, while $NN_{H_1}$ uses a linear activation function. The learning rate is set to $0.001$, while the batch size is fixed with a value of 10. The maximum number of epochs used in this case is $15,000$. No regularization is used while training the different ML models.

\subsubsection{Results}
\begin{figure}[t!]
    \centering
        \begin{subfigure}[]{0.49\textwidth}
            \includegraphics[width=\textwidth]{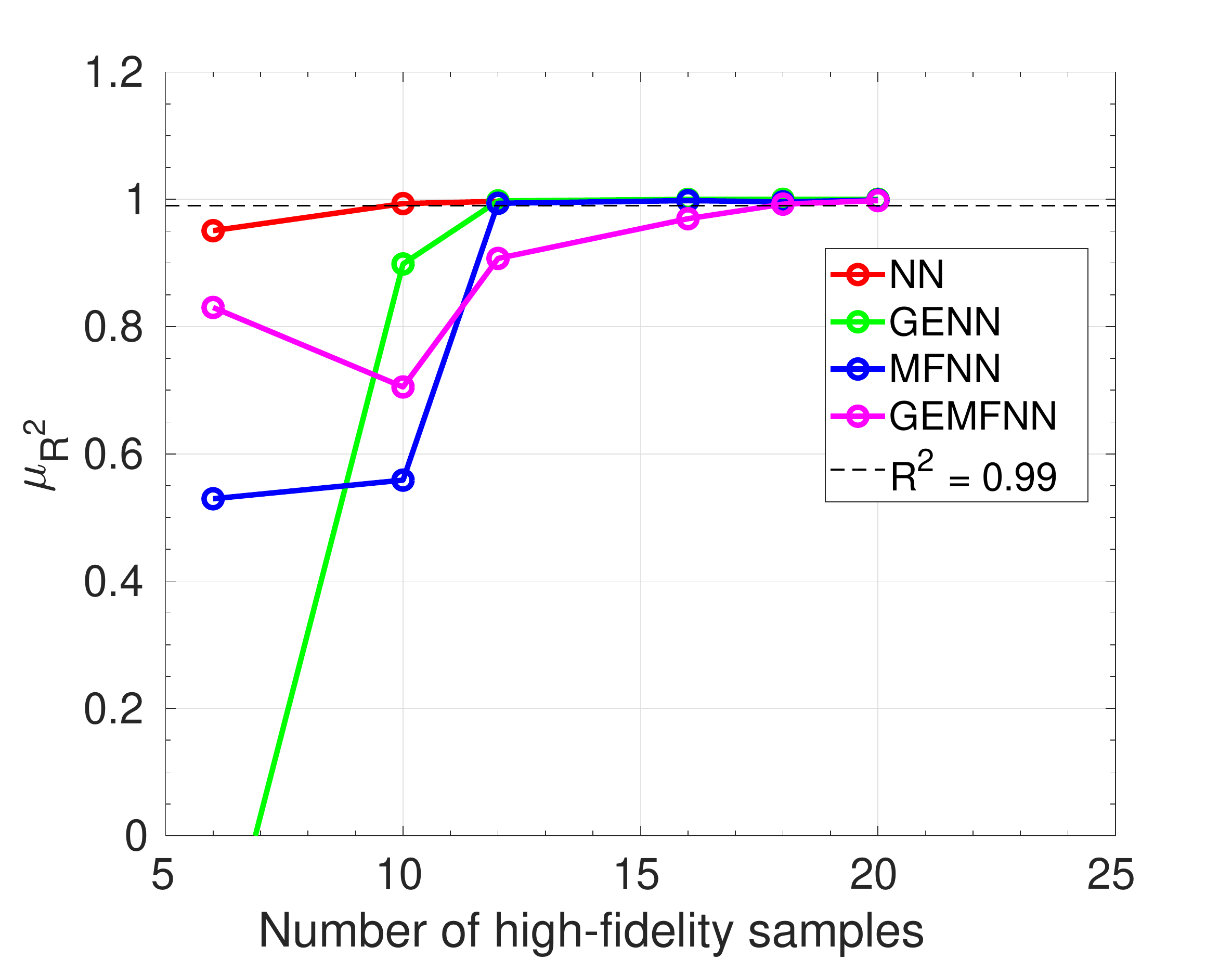}
            \caption{}
        \end{subfigure}
        \begin{subfigure}[]{0.49\textwidth}
            \includegraphics[width=\textwidth]{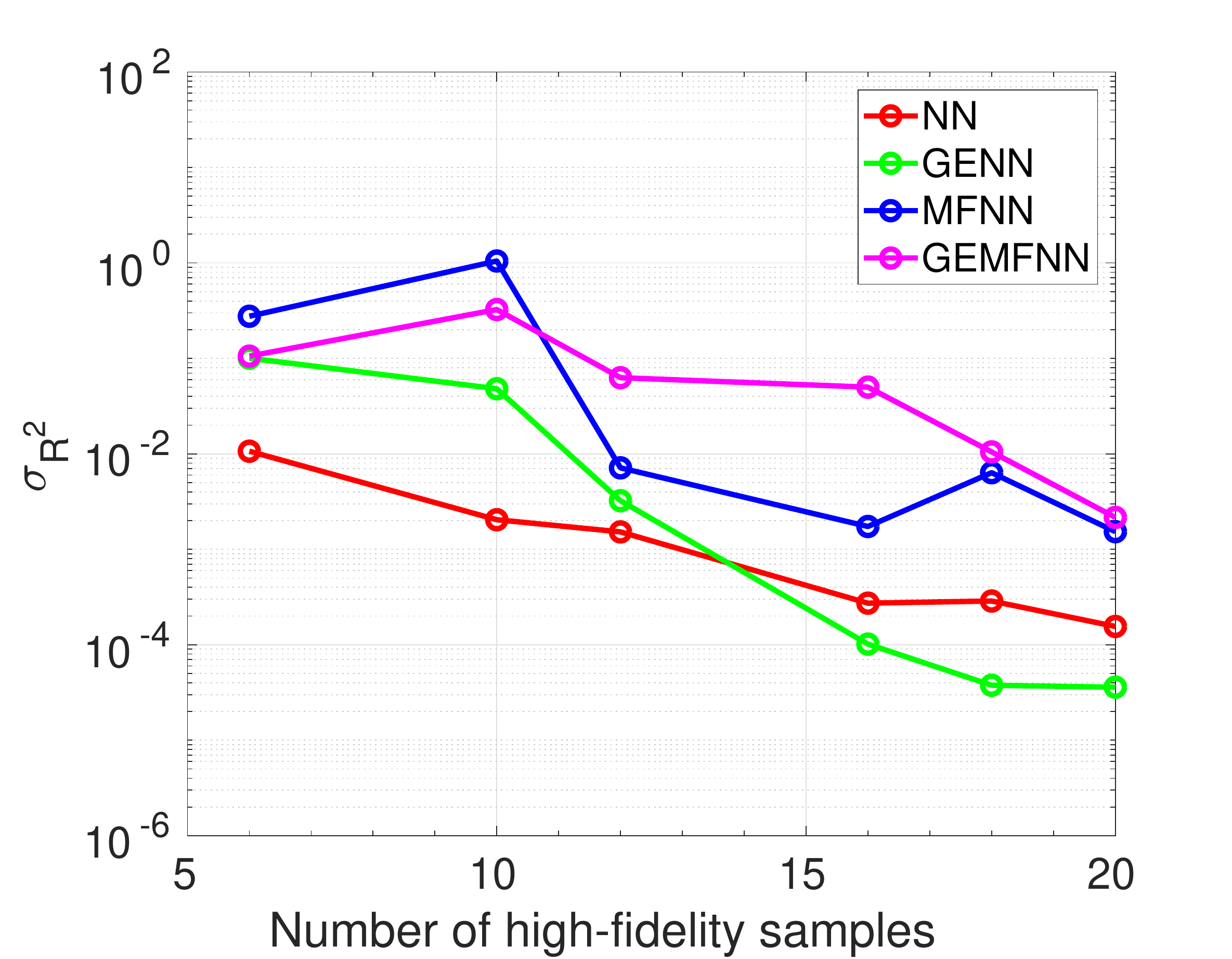}
            \caption{}
        \end{subfigure}
    \caption{One dimensional function results: (a) mean of $R^2$, (b) standard deviation of $R^2$.}
    \label{r2_1}
\end{figure}

Table~\ref{table_1d} shows the number of high-fidelity samples required by each ML model to reach the global accuracy of $R^2 = 0.99$. The multifidelity models use an additional of 500 low-fidelity sample points. Note that the number of sampling points for the gradient-enhanced cases accounts for both the cost of evaluating the function and its gradient. Therefore, $200$ sampling points for these cases, for example, correspond to $100$ function and $100$ gradient evaluations. Figure \ref{r2_1}(a) shows that NNs outperformed all the other models, requiring only ten high-fidelity samples to reach the global accuracy. GENNs and MFNNs both required twelve samples each to reach this accuracy, while GEMFNNs required $18$ samples. The results in Fig. \ref{r2_1}(a) are generated by averaging the outcomes from ten different datasets. The corresponding standard deviation for each model is shown in Fig. \ref{r2_1}(b). Figure \ref{r2_1}(b) shows that the standard deviation decrease with increase number of samples, resulting in models that are less sensitive to the training data, as well as due to the stochastic nature of the ADAM optimizer \cite{kingma2014}. For this case, the cost of evaluating the low-fidelity model is neglected. The results for this case imply that there is no benefit of using gradients as well as data from different levels of fidelity in NNs when the model is of a low-dimensional input space.  

\begin{table}[t]
\caption{One dimensional function modeling cost.}
\begin{center}
\label{table_1d}
\begin{tabular}{c l l}
& &\\ 
\hline
ML model& Modeling cost\\
\hline
NN & 10\\
GENN & 12$^*$\\
MFNN & 12$^{**}$\\
GEMFNN & 18$^{*,**}$\\
\hline
$^*$Function plus gradient evaluation cost\\
$^{**}$Plus 50 low-fidelity training samples
\end{tabular}
\end{center}
\end{table}

\subsection{Case 2: two dimensional Rastrigin function}
The two dimensional Rastrigin function \cite{Rastrigin1974} used in this study is written as
\begin{equation}
    f_{\text{HF}}(\textbf{x}) = 20 + \sum_{i=1}^{2}(x_i^2 - 10cos(2\pi x_i)),
\end{equation}
where $\textbf{x}$ $\in$ [-1,1.5]. For this study the low-fidelity model

\begin{equation}
    f_{\text{LF}}(\textbf{x}) = 0.5f_{\text{HF}}(\textbf{x}) + \sum_{i=1}^{2}(x_i-0.5 )
\end{equation}
is introduced. The corresponding high-fidelity model gradient is
\begin{equation}
    \nabla f_{\text{HF},i}(\textbf{x}) = 2x_i + 20\pi sin(2\pi x_i),
\end{equation}
where $i=1,2$, and the low-fidelity model gradient is
\begin{equation}
    \nabla f_{\text{LF},i}(\textbf{x}) = 0.5\nabla f_{\text{HF},i}(\textbf{x}) + 1.
\end{equation}

\begin{table}[t!]
\caption{Rastrigin function modeling cost.}
\begin{center}
\label{table_2d}
\begin{tabular}{c l l}
& &\\ 
\hline
ML model& Modeling cost\\
\hline
NN & 150\\
GENN & 120$^*$\\
MFNN & 150$^{**}$\\
GEMFNN & 120$^{*,**}$\\
\hline
$^*$Function plus gradient evaluation cost\\
$^{**}$Plus 500 low-fidelity training samples
\end{tabular}
\end{center}
\end{table}

\begin{figure}[t!]
    \centering
        \begin{subfigure}[]{0.49\textwidth}
            \includegraphics[width=\textwidth]{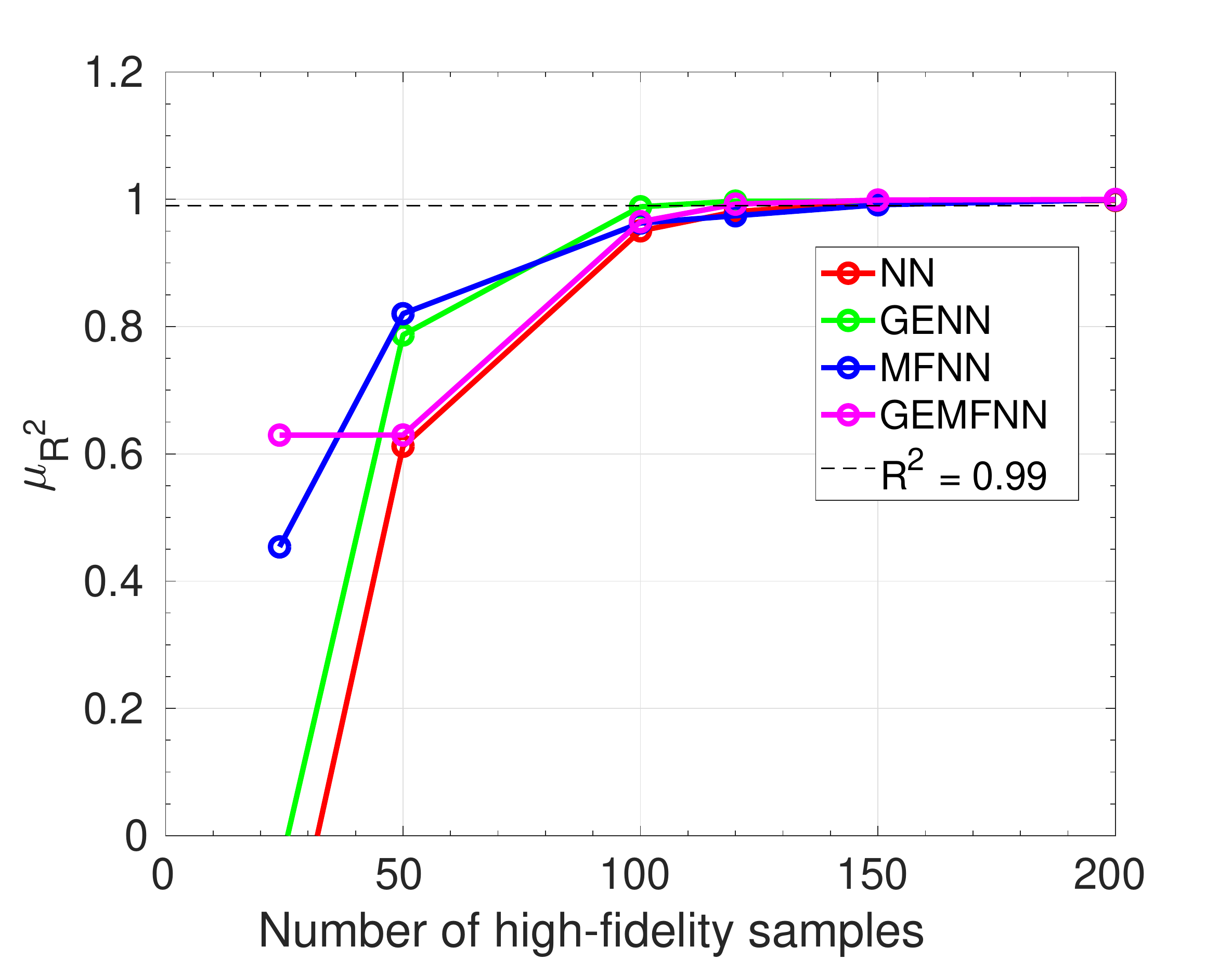}
            \caption{}
        \end{subfigure}
        \begin{subfigure}[]{0.49\textwidth}
            \includegraphics[width=\textwidth]{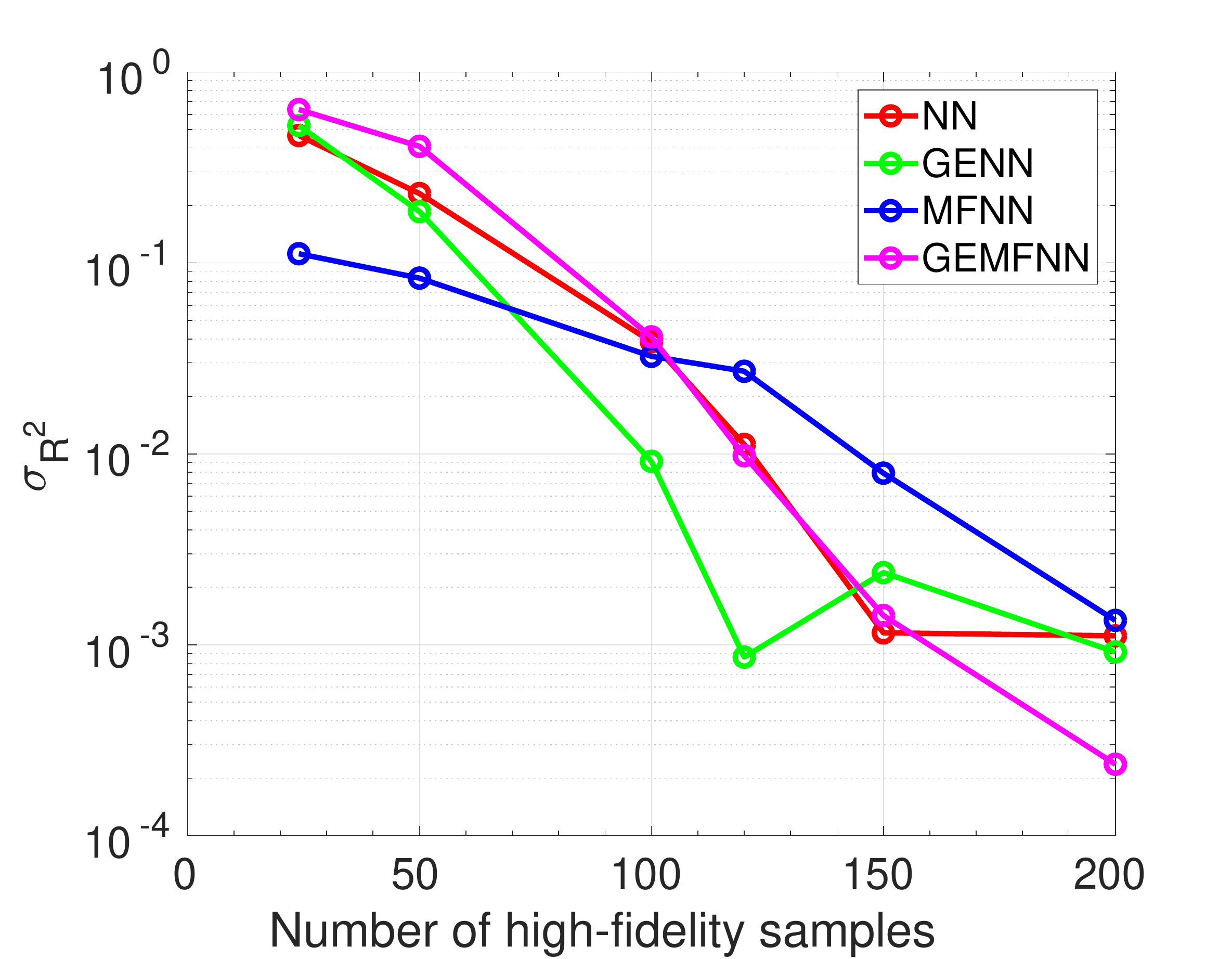}
            \caption{}
        \end{subfigure}
    \caption{Rastrigin function results: (a) mean of $R^2$, (b) standard deviation of $R^2$.}
    \label{r2_2}
\end{figure}

\subsubsection{ML model setup}
The LHS plan is used to generate the low- and high-fidelity training data, while the full factorial sampling plan is used to generate the testing data. The testing dataset consists of $10,000$ samples. $NN_{L}$ and $NN_{H_2}$ use two hidden layers, with $50$ neurons each, while $NN_{H_1}$ contains only one hidden layer with ten neurons. Similar to the previous case, $NN_{L}$ and $NN_{H_2}$ use the tangent hyperbolic activation function, and $NN_{H_1}$ contains a linear activation function. The batch size, the learning rate and total number epochs used in this study are set to $32$, $0.001$, and $10,000$, respectively. No regularization was used.

\subsubsection{Results}
The number of high-fidelity samples required to reach the global accuracy threshold of $R^2 = 0.99$ is shown in Table \ref{table_2d}. The multifidelity models also use an additional of $500$ low-fidelity samples during its construction. The variation of the $R^2$ error metric with number of high-fidelity samples is shown in Fig. \ref{r2_2}(a). The results in Fig. \ref{r2_2}(a) are averaged over ten different datasets. Figure \ref{r2_2}(b) shows the corresponding standard deviations for the same datasets. GENNs and GEMFNNs both require $120$ high-fidelity sample points to reach the target threshold. NNs and MFNNs, other the other hand, require $150$ high-fidelity samples to reach the same threshold. Similar to the previous case, the standard deviation of the $R^2$ metric decrease with increasing sample size, as seen in Fig. \ref{r2_2}(b). For this case, the use of multifidelity data does not add to an improvement in predictive capabilities of the ML models, however, using gradient information does. GENNs, hence, work best for modeling the Rastrigin function.

\subsection{Case 3: 20 dimensional analytical function}

The high-fidelity model of the 20 dimensional analytical function \cite{Meng2020} is written as
\begin{equation}
    f_{\text{HF}}(\textbf{x}) = (x_1-1)^2 + \sum_{i=2}^{20}(2x_i^2 - x_{i-1})^2,
\end{equation}
where $\textbf{x}$ $\in$ [-3,3]. The corresponding low-fidelity model is \cite{Meng2020}
\begin{equation}
    f_{\text{LF}}(\textbf{x}) = 0.8f_{\text{HF}}(\textbf{x}) - \sum_{i=1}^{19}0.4x_ix_{i+1} - 50.
\end{equation}
The high-fidelity gradient is
\begin{equation}
    \nabla f_{\text{HF},1}(\textbf{x}) = 2(x_1-1) - 2(2x_2^2 - x_1),
\end{equation}
\begin{equation}
    \nabla f_{\text{HF},i}(\textbf{x}) = 8x_i(2x_i^2 - x_{i-1}) - 2(2x_{i+1}^2 - x_i), 
\end{equation}
for $i = 2,3,...,19$, and 
\begin{equation}
    \nabla f_{\text{HF},20}(\textbf{x}) = 8x_{20}(2x_{20}^2 - x_{19}).
\end{equation}
The low-fidelity gradient is
\begin{equation}
    \nabla f_{\text{LF},1}(\textbf{x}) = 0.8\nabla f_{\text{HF},1}(\textbf{x}) - 0.4x_2,
\end{equation}

\begin{equation}
    \nabla f_{\text{LF},i}(\textbf{x}) = 0.8\nabla f_{\text{HF},i}(\textbf{x}) - 0.4(x_{i-1} + x_{i+1}),
\end{equation}
for $i = 2,3,...,19$, and
\begin{equation}
    \nabla f_{\text{LF},20}(\textbf{x}) = 0.8\nabla f_{\text{HF},20}(\textbf{x}) - 0.4x_{19}.
\end{equation}

\begin{table}[b!]
\caption{20 dimensional function modeling cost.}
\begin{center}
\label{table_20d}
\begin{tabular}{c l l}
& &\\ 
\hline
ML model& HF sample cost\\
\hline
NN & $\textgreater$10,000\\
GENN & 5,000$^*$\\
MFNN & $\textgreater$10,000$^{**}$\\
GEMFNN & 600$^{*,**}$\\
\hline
$^*$Function plus gradient evaluation cost\\
$^{**}$Plus 30,000 low-fidelity training samples
\end{tabular}
\end{center}
\end{table}

\begin{figure}[b!]
    \centering
        \begin{subfigure}[]{0.49\textwidth}
            \includegraphics[width=\textwidth]{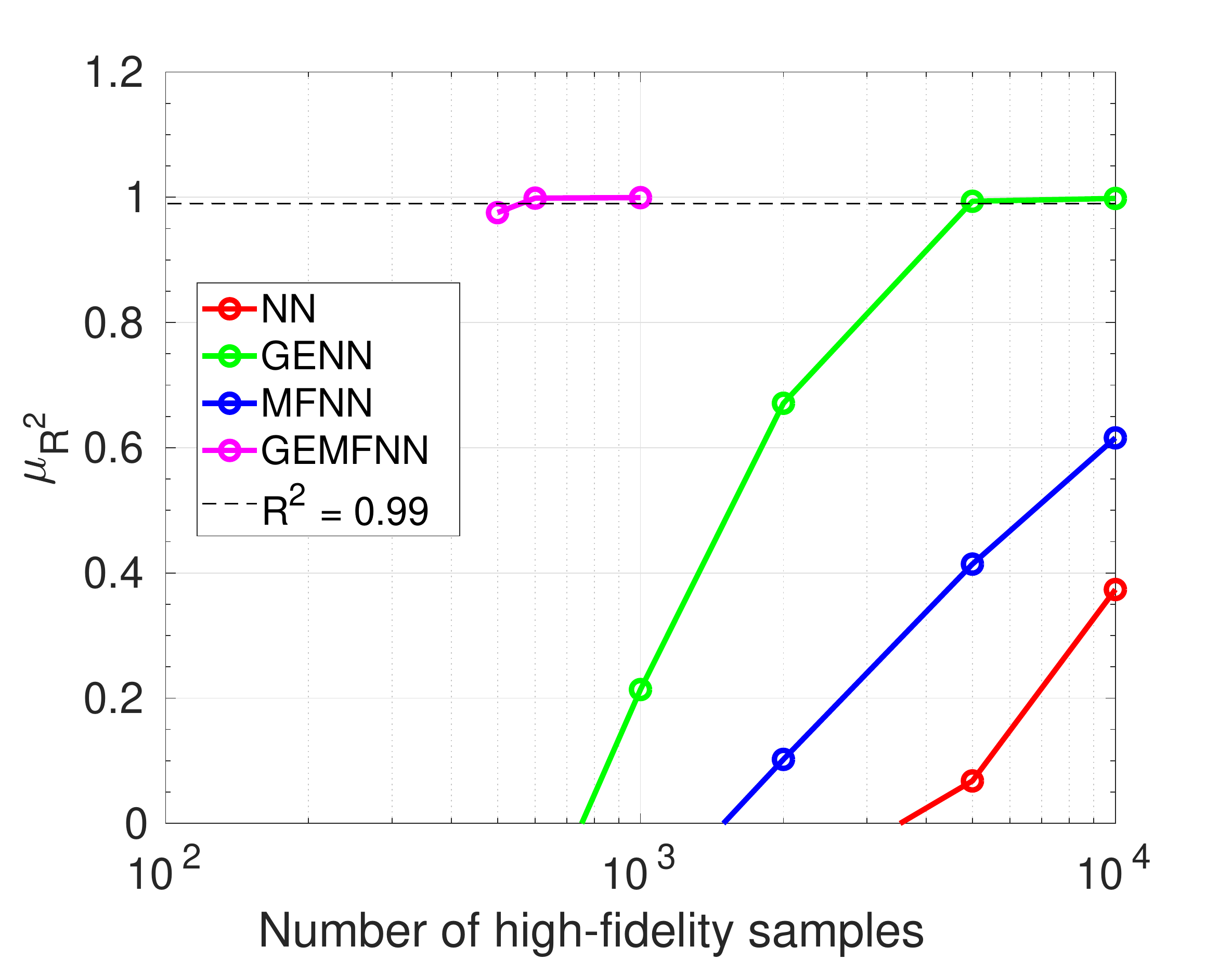}
            \caption{}
        \end{subfigure}
        \begin{subfigure}[]{0.49\textwidth}
            \includegraphics[width=\textwidth]{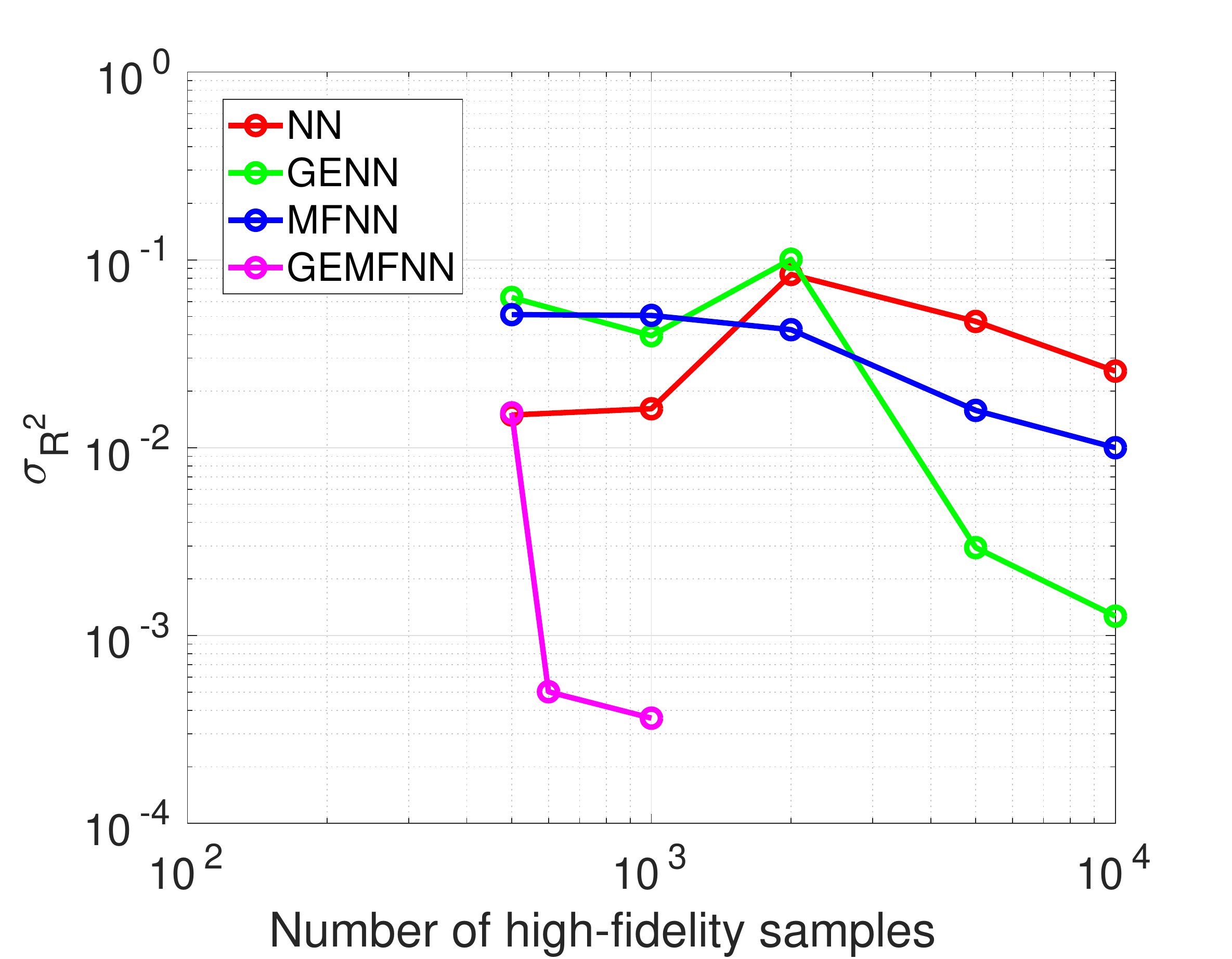}
            \caption{}
        \end{subfigure}
    \caption{20 dimensional function results: (a) mean of $R^2$, (b) standard deviation of $R^2$.}
    \label{r2_3}
\end{figure}

\subsubsection{ML model setup}
For this case, the LHS plan was used to generate the training and testing data. The testing data contains $10,000$ samples. Similar to the previous two cases, $NN_{H_1}$ contains only one hidden layer with ten neurons. $NN_{H_2}$ contains four hidden layers, with $64$ neurons, while $NN_{L}$ contains six hidden layers with $128$ neurons. No regularization was used, while the batch size, learning rate, and number of epochs were set to $64$, $0.001$, and $10,000$, respectively.

\subsubsection{Results}
Table \ref{table_20d} shows the high-fidelity sample cost required by each ML model to reach the target accuracy. GEMFNNs far outperforms the other models and requires around eight times fewer samples compared to its nearest competitor, GENNs, which requires $5,000$ samples. Both NNs and MFNNs fail to meet the target accuracy, even with $10,000$ samples. Figures \ref{r2_3}(a) and \ref{r2_3}(b) show the mean and standard deviations, respectively, of $R^2$ with respect to the number of high-fidelity samples, performed using ten different datasets. The mean of $R^2$ increases, while the standard deviation of $R^2$ decreases with increasing sample sizes. The benefit of both gradient and multifidelity information in training the NNs is demonstrated for a high-dimensional problem and is shown in this case.

\section{Conclusion}
The GEMFNNs ML model is applied to three analytical cases, namely, a one, two and a 20 dimensional variable problem, and is compared to NNs, GENNs, and MFNNs. GEMFNNs are a multifidelity version of GENNs and its construction is similar to MFNNs. GEMFNNs consists of three NNs, one to approximate the low-fidelity data ($NN_L$), which is then connected to two other NNs, one with linear ($NN_{H_1}$) and the other with nonlinear ($NN_{H_2}$) activation functions, in order to capture both linear and nonlinear correlations, respectively, between high- and low-fidelity data. In NNs, the loss function used is the MSE between the true and predicted function values. In GENNs, this loss function is modified by adding the MSE of the true and predicted gradient values to the original loss function. 

NNs outperformed all the other models in the one dimensional case, while GENNs did the same for the two dimensional case and GEMFNNs for the 20 variable case. This study shows the benefit of using both gradient and multifidelity information in training the ML models for high-dimensional cases. It also shows, that for low dimensional cases, using multifidelity information does not improve predictive performance of the ML models. 

In this study, for each case, the same hyperparameters are used for all the models. This may not be ideal for individual models as they might perform better with different hyperparameters. Using different hyperparameters such as different activation functions and the addition of regularization will need to be done.

This study has been conducted using analytical benchmark cases. While they do not represent engineering problems, they are a good starting point in testing ML models. Future applications of the above models will done on problems involving optimum design, uncertainty quantification and global sensitivity analysis.

\bibliographystyle{unsrt}  
\bibliography{references}  






\end{document}